\documentclass[a4paper,10pt]{article}
\pdfoutput=1
\usepackage[utf8]{inputenc}
\usepackage{cite}
\usepackage{amsmath,amssymb,amsfonts}
\usepackage{algorithmic}
\usepackage{graphicx}
\usepackage{textcomp}
\usepackage{xcolor}
\usepackage{tikz}
\usetikzlibrary{positioning, shapes}
\usetikzlibrary{calc}

\title{\LARGE \bf Theory of Mind Based Assistive Communication in Complex Human Robot Cooperation\\}

\author{Moritz C. Buehler$^{1, 2}$, J\"urgen Adamy$^{1}$ and Thomas H. Weisswange$^{2}$\\
\thanks{$^{1}$ with Control Methods \& Robotics Lab, Technical University Darmstadt, Germany}
\thanks{$^{2}$ with Honda Research Institute Europe GmbH, Germany}
}

\begin{document}

\maketitle

\begin{abstract}
When cooperating with a human, a robot should not only care about its environment and task but also develop an understanding of the partner's reasoning. 
To support its human partner in complex tasks, the robot can share information that it knows. However simply communicating everything will annoy and distract humans since they might already be aware of and not all information is relevant in the current situation. 
The questions when and what type of information the human needs, are addressed through the concept of Theory of Mind based Communication which selects information sharing actions based on evaluation of relevance and an estimation of human beliefs. 
We integrate this into a communication assistant to support humans in a cooperative setting and evaluate performance benefits. 

We designed a human robot Sushi making task that is challenging for the human and generates different situations where humans are unaware and communication could be beneficial. We evaluate the influence of the human centric communication concept on performance with a user study. Compared to the condition without information exchange, assisted participants can recover from unawareness much earlier. The approach respects the costs of communication and balances interruptions better than other approaches. By providing information adapted to specific situations, the robot does not instruct but enable the human to make good decision.  
\end{abstract}

\section{Introduction}

In complex environments, such as driving scenarios, navigation tasks or emergency rescue coordination, many different types of information are important to handle a situation appropriately. To maintain situation awareness for environment and task, information has to be perceived, combined and processed, and finally used to anticipate the evolution of the current situation \cite{Endsley1995}. 

With the improvement of technical systems like robots and AI, more interest shifts towards interacting with and assisting humans in complex environments where many aspects need to be taken into account to achieve and maintain situation awareness. 
There exist many specific assistance approaches designed for single and limited use cases. 
Some approaches take a very direct approach, telling the human what to do which might prevent immediate errors effectively. However, to help the human to regain awareness and to enable good future decisions it might be better to analyze the problem and provide necessary pieces of information to the human.

As an example, a blind spot assistant in a car encodes an implicit understanding of the human behavior. It is based on the static perception model, that the human cannot observe objects in the car's blind spot area. Based on an erroneous belief of a free left lane, the driver might decide to indicate lane change and initiate a steering maneuver. With this build-in understanding, the assistance system can successfully intervene and signal the presence of another car through visual hints and audio warning, thereby allowing humans to correct their decision.

Interrupting a human is not always necessary and not all information is relevant in every situation. Instead every interaction with the human requires attention, can distract from or delay other important tasks and annoy the human. As long as a driver does not intent to change lanes, the presence of a car on the other lanes is not relevant. This changes, when the driver indicates lane change and the more disturbing auditory warnings are justified. 

A blind spot assistant implicitly encodes a human model and relevance assessment for its specific use case. 
When respecting different types of information, human knowledge and relevance further relate to the question what to communicate.

For a more general support to a human partner, a robot assistant will have to reason in a similar way to decide, when and what type of information to share, to flexibly support a human partner according to her needs. 
To enable such human centric support in different situations (not limited to the blind spot case), an explicit understanding of task and human mental processes, a theory of mind (ToM), is helpful to flexibly detect and evaluate current communication needs.

ToM describes the inference of others mental states such as beliefs, desires or intentions \cite{Wimmer1983}. 
The development of a theory of mind is an important capability for humans and a basis for human interaction and communication \cite{Happe1993}. 

We previously introduced an artificial theory of mind that allows us to develop an understanding of the human during interaction \cite{Buehler2018}. It can be combined with a task model to evaluate relevance of information and rate it in respect to communication costs in an explicit and generic way. This was included in our receiver centric communication concept of theory of mind based communication which we formalized in prior work for abstract POMDPs \cite{Buehler2020}. This concept considers what humans might already know and what they need to know to handle the current situation appropriately, leading to the decision when and what type of information should be shared to support the receiver's awareness. 
In this paper, we present an information sharing assistant integrating this communication concept to support humans in a challenging domain. We propose a complex human robot cooperative task, to investigate and evaluate effectiveness of the theoretical concept in the interaction with human participants.

\begin{figure}
\centering
\includegraphics[width=.95\linewidth]{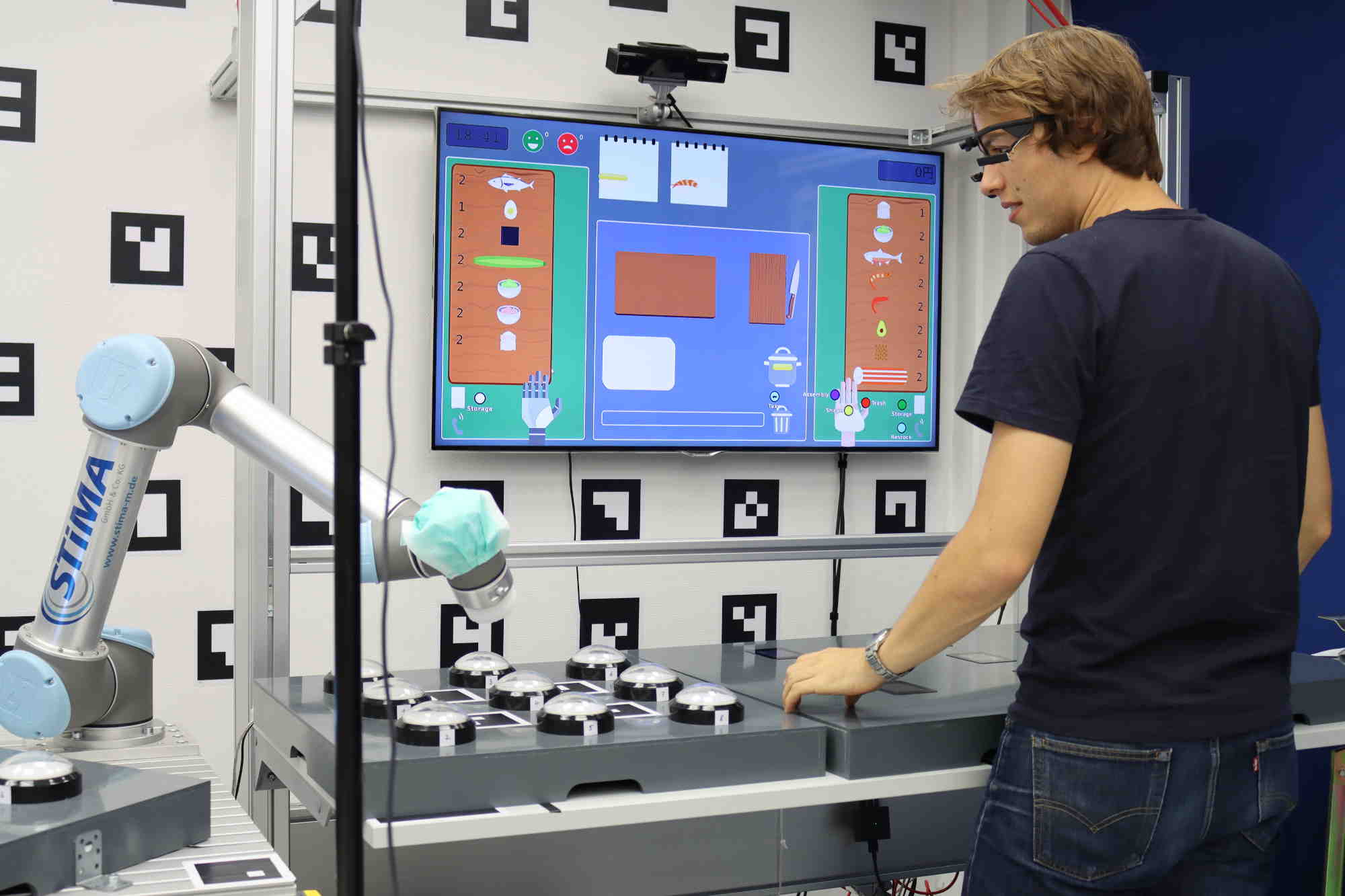}
\caption{Human robot cooperative task setup.}
\label{setup}
\end{figure}
 
 \section{Related work}
Our human centric communication concept can be divided into two parts, human understanding and the evaluation of communication influence. By interpreting observed human behavior, we estimate information aspects the human might have missed. The evaluation of their importance for the task leads to the decision when and what type of information should be shared. 
The question how communication should be designed is not in our focus here, explicit as well as implicit communication forms have been considered in previous human robot interaction works \cite{Torrey2013}, \cite{Dragan2014}, \cite{Reddy2020}. 

To support a human in one specific decision, classical assistance approaches compare human behavior to some predefined or optimal reference. When a deviation is detected, a correcting action is proposed. This concept was for example applied to healthcare and automotive contexts to prevent human errors \cite{Hoey2013}, \cite{McCall2007}. 

More advanced methods require an explicit understanding of human behavior. 
One principled approach to understand hidden causes of human behavior such as intentions and beliefs is to collect human trajectories and search for explanations in retrospective. Inverse Reinforcement Learning \cite{Ng2000} estimates reward functions that can explain human actions. 
Baker et al. \cite{Baker2017} extend the estimation of human intentions to also account for human state uncertainty. They introduce a perception model and develop a Bayesian observer approach to estimate human desires and beliefs.
In repetitive or static settings such estimated rewards or beliefs can be used to predict human behavior in reoccuring situations \cite{Sadigh2016}, \cite{Ferreira2015}.

During interaction human mental states may change and it can be necessary to infer them continuously. 
In some situations a single hidden state is relevant to coordinate behaviors. 
Unhelkar et al., \cite{Unhelkar2020} estimate the intended next human action based on observations to resolve collisions. Nikolaidis et al., \cite{Nikolaidis2018} infer human's willingness to adapt to a robot's plan based on their reactions when moving objects together.

The relevance of different information types can change during a task as well. 
Task relevance is a key aspect in critical situations or in multi agent exploration settings, where others beliefs are known (for example as they have access to non-overlapping parts of the observation space). Here, relevance can be evaluated by the influence of information sharing on other agents belief and expected behavior. This can be used to decide when and what type of information is shared between agents \cite{Goldman2003}, \cite{Melo2011}, \cite{Roth2006}. 
The relevance can further depend on static preferences of a receiver \cite{Chitnis2018} or the reliability and exclusiveness of the communication channel \cite{Renoux2020}.

When collocated, a robot can observe the human and interpret action and information gathering and the task relevance of information should be combined with an understanding of the human. With the overlap in perception, the estimation of other's knowledge becomes important to avoid sharing information that the other already is aware of. In previous work, we proposed inference of a multi dimensional human belief during interaction based on a model of human perception and decision making \cite{Buehler2018}. The influence of deviations in the belief were used to assess human situation awareness. 
We further introduced a conceptual extension that explicitly incorporates the expected benefit of communication and its cost into the model \cite{Buehler2020}. In this paper, we integrate this approach into an assistive communication module to support a human in a real collaboration with a robot.

\section{Theory of mind based communication}

We will shortly review the general concept of theory of mind based communication (ToM-Com) that was introduced in \cite{Buehler2020}. This will be the basis for the robot's communicative assistance module. 

To account for a human partner that may not represent every important environmental information correctly, 
we model her as an actor in a Partially Observable Markov Decision Process (POMDP) \cite{Kaelbling1998}. 
While not having access to the true state at all time, it only receives observations (e.g.\ through visual perception), that may reveal some aspects of the true state according to an observation function $O_H(s, o_H) = p(o_H\vert s)$. The human can maintain an internal representation of the environment, the belief, as a probability distribution over states, $b_H = p(s)$. 
On receiving observations, it can update the belief as a Bayesian observer according to the observation likelihood, \begin{align}
p(s\vert o_H) \sim p(o_H \vert s )p(s).
\label{observation}
\end{align}

The human can influence the state transition through his action $a_H$. 
Note that the robot's actions are not explicitly modeled here, but included into the state transition function \begin{align}
T(s, a_H, s') = \sum_{a_R}p(s'\vert s, a_H, a_R) p(a_R).
\label{transition}
\end{align}

The state $s$ factors in different state aspects $s_i$ corresponding e.g. to the substate of different locations. 
The cooperative task goal is encoded with a reward function $R_H(s, a_H, s')$. 
The human decision making is modeled to be noisy rational according to a softmax policy \begin{align}
p(a_H\vert b_H)\sim \exp(\tau Q_H(b_H, a_H)), 
\label{policy}
\end{align} where $Q_H(b_H, a_H)$ describes the action value function that is based on the current human belief $b_H$. This stochastic policy specifically accounts for noise respectively unpredicted effects in human decision making or action execution.

\subsection{Inference of human belief} 
The first part of our concept of theory of mind based communication consists in interpreting human behavior to estimate mental states. 
The human model is used to infer the human belief at any time by observing human decisions and information gathering.

This second level inference structure, inference of human state inference, is computationally not tractable for large state spaces. Therefore, we constrain the representation of the inferred belief. We approximate the belief by factorizing distributions according to the independent state aspects $s_i$ and use a Dirichlet distribution for each factor, $p(b_H) \approx \prod_i Dir(b_{H, i} \alpha_i)$. 

By observing the human we can predict the belief changes according to the human perception model eq.\ (\ref{observation}) and state transition eq.\ (\ref{transition}). The human action provides feedback from human decision making eq.\ (\ref{policy}) and is used to update the belief according to $p(b_H\vert a_H)\sim p(a_H\vert b_H)p(b_H)$. 
For these steps we sample the prior human belief estimate, apply the updates and remap the resulting distribution back to the Dirichlet distributions by moment matching \cite{Minka2000}.

\subsection{Planning Communication}
To decide what and when to communicate to support the human, the robot evaluates relevance by estimating communication benefits and comparing it to the cost of communication. 

The decision for communication actions $a_\mathrm{comm}$ results from a second POMDP which includes the human belief as part of the robot's state space, $s_R = (s, b_H, a_H)$. Here we assume full task knowledge, the robot can observe the true state and the human action. 
Regarding relevance, the robot needs to estimate communication effects on human beliefs which in consequence may change human behavior (eq. (\ref{policy})).

The transition function for the communication POMDP combines the human model eqs.\ (\ref{observation})-(\ref{policy}), and the influence of communication on the human belief, 
\begin{align}
p(s_R'\vert s_R, a_R) &= p(s', b_H', a_H' \vert s, b_H, a_H, a_R) \nonumber\\&= 
	\sum_{o_H', b_{H-}}\Bigl(\underbrace{p(b_H'\vert a_H, b_{H-}, o_H')}_{\text{belief update}}\underbrace{p(o_H' \vert s')}_{\text{human perception} }\nonumber\\&
	\underbrace{p(s' \vert s, a_H')}_\text{state trans.} \underbrace{p(a_H' \vert b_{H-})}_{\text{human decision}}\underbrace{p(b_{H-}\vert a_R, b_H)}_{\text{comm. effect}}\Bigr).
\end{align}
An information sharing action updates the human belief corresponding to the communicated information and might help the human to chose better actions. To account for distraction or time delay, a cost of communication $-R_\mathrm{comm}(s, a_\mathrm{comm})$ is included in the reward function of the robot $R_R = R_H + R_\mathrm{comm}$. Hence, the planning process balances the expected benefits due to improved human decision making against the distraction or other negative effects of communication represented in the immediate reward to select when and what information to communicate to the human.

\section{Evaluation}

We evaluate the interaction of a robot equipped with a ToM-Com assistance module cooperating with human participants.

\subsection{Human robot cooperative task}
\paragraph{Requirements}
Many tasks for human robot interaction are relatively easy to solve for human participants, since they are designed with a main focus on the robot (e.g.\ the tasks in \cite{Unhelkar2020}, \cite{Nikolaidis2018}). To evaluate quantitative effects of human support, however, we require a task that challenges the human. This task should generate frequent situations where a human participant is not aware of some aspect of the situation and may profit from communicated information. Such situations are necessary to evaluate the benefits of our principled communication approach, however our approach is not limited to such setups (indeed, many warning systems are constructed for rare but severe events). 
We consider the questions what and when to share information to support a human receiver. 
To require solving the problem of what to communicate, the task should contain potential error sources that can be resolved by communicating different types of information. 

Support of a human in dynamic situations is only possible if human awareness problems can be detected from available observations during interaction time (and not only retrospectively), which means there needs to be some indication. Often, a problem in human awareness influences their behavior over a period of time where the first time steps can be sufficient to detect future problems.

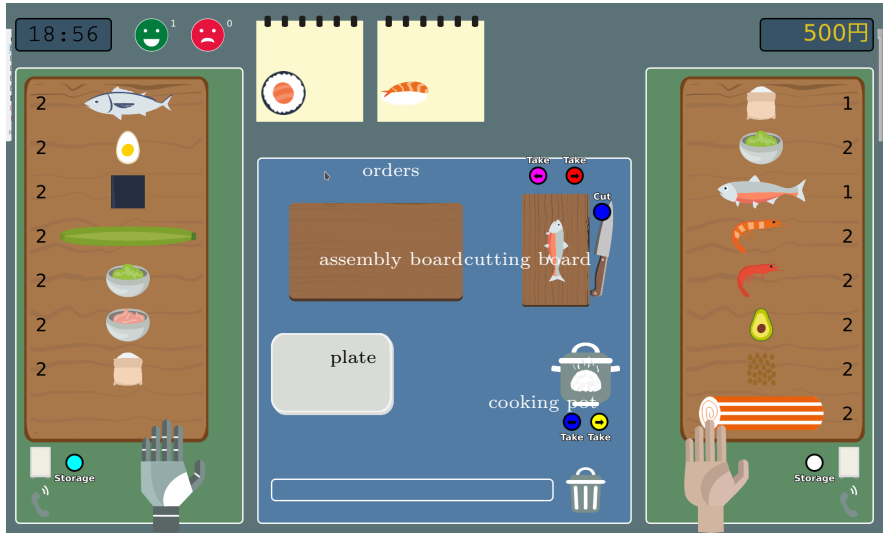
\begin{figure}
\centering

\begin{tikzpicture}
  	\node[anchor=center] at (0,0) {
      \includegraphics[width=.95\linewidth]{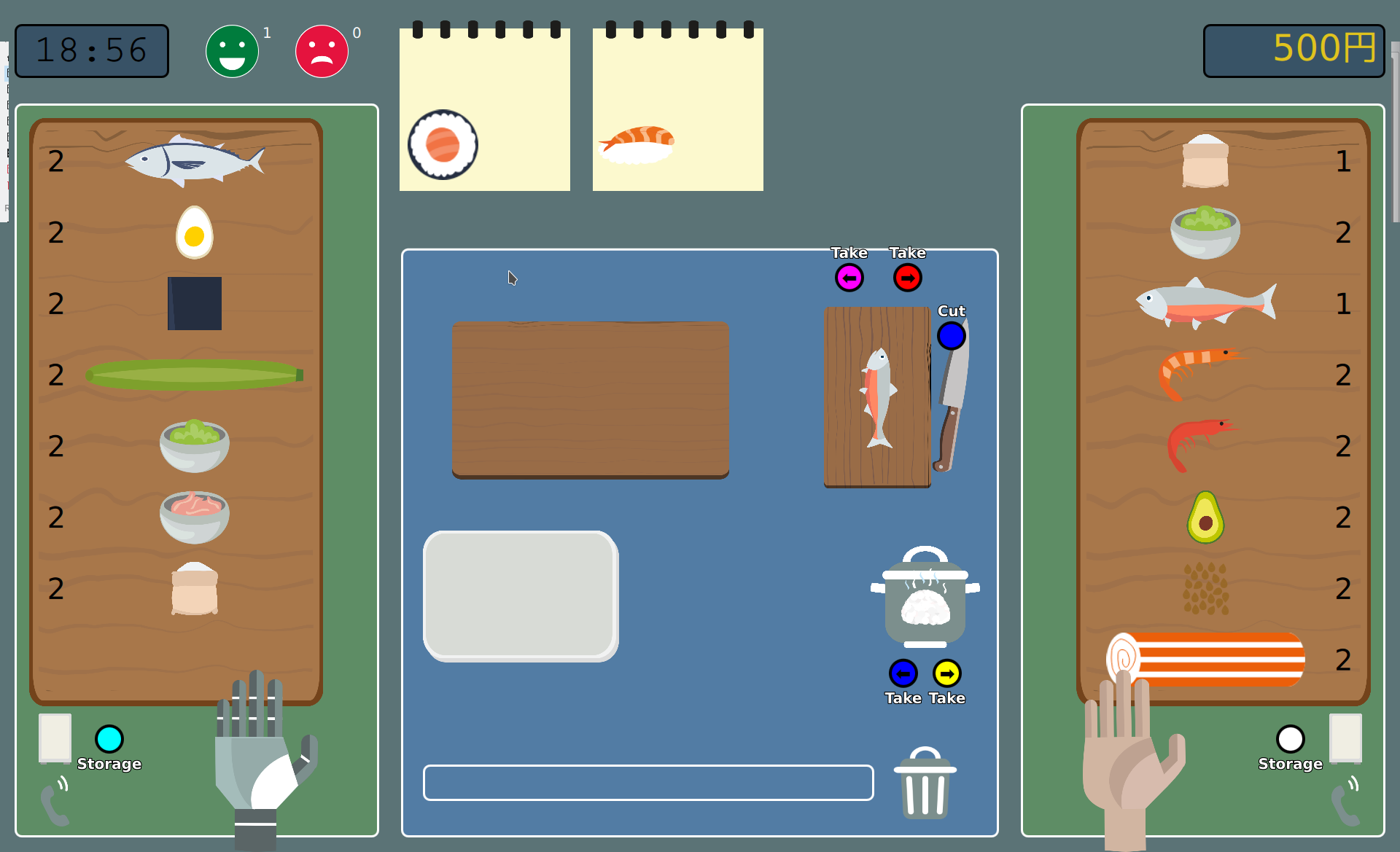}};
      \node (orders)[ text=white] at (-.7, 1.3){\scriptsize orders};
      \node (assembly)[text=white] at (-.7, .1) {\scriptsize assembly board};
      \node [text=white] at ($(assembly)+(1.8, 0)$) {\scriptsize cutting board};
      \node [text=black] at ($(assembly)+(-.5, -1.3)$) {\scriptsize plate};
      \node [text=white] at ($(1.3, -1.8)$) {\scriptsize cooking pot};
      \end{tikzpicture}

\caption{Sushi environment visualization. Robot area on the left, human area on the right and shared processing locations in the center. Customer orders are presented at the top.}
\label{gui}
\end{figure}
\paragraph{Sushi task}
We developed a cooperative sushi making task. A robot and a human have to assemble different Sushis cooperatively according to customer orders (see Fig. \ref{setup} for our task setup). 

The task concept is similar to the video game Overcooked\footnote{{https://www.team17.com/games/overcooked/}} in that it challenges human situation awareness and coordination in a dynamic kitchen environment. 

Overcooked was previously proposed to evaluate human interaction effects \cite{Bishop2020}. 
We explicitly add a physical interface for human robot cooperation.

Human and robot both have access to an array of basic ingredients. These have to be processed to assemble different sushi recipes and serve customer orders. 

Some ingredients need to be cut at the cutting board, cooked at the cooking pot, shaped by the human hand, and then finally stacked together at the assembly board. 
About 15 subsequent actions are necessary until an order can be served from the plate. If something went wrong, unneeded ingredients can be trashed. 
The actions of the task are mapped to physical buttons which can be pressed by the agents. Correspondence is defined through colors and areas of actions and buttons. 

We use 6 different Sushi recipes to achieve an adequate level of difficulty. 
Two customer orders are generated with random content. When one order is served to the customer, a new one appears after a short delay. The orders do not show the full recipe but only the final product. 
Except for the assembly board, each location can only carry one item, which challenges planning and coordination.

The task visualization (see Fig. \ref{gui}) is presented on a large display on the wall. 
We use a UR5 robot, constructed for operating with humans in a common workspace. The robot stops when a collision is detected or its intended action executed. 
To measure human information gathering, we use eye tracking glasses from PupilLabs to extract human eye gaze and a marker based mapping to world coordinates. An area around the gaze fixation is considered to generate observations and each gaze sample leads to a corresponding observation update. Since the availability of actions depends on the location states, this can serve as additional observation for the human.

\paragraph{Task complexity}
We designed the task to be difficult for the human to handle. One type of challenges for the human is induced by the design of ingredients, we e.g. use different similar looking fish and shellfish. The necessary actions are changing due to random appearance of Sushi orders which require different ingredients and also different processing steps. 
The cooperation with the robot provides the chance to distribute work between agents but this needs to be coordinated. Coordination is necessary on the strategic level (which order to prepare first) as well as on the action level (who contributes which ingredient). 
Action planning is difficult due to the complexity of the recipes and the restriction that most locations can only hold one item. 
So we challenge the human in different regards, memorizing different recipes, differentiating the ingredient symbols, planning actions, and coordinate with the robot partner. 

\paragraph{Belief representation}

As states represented in the human belief model, we consider the location contents and recipe modifications. For each location (besides ingredient storage) we introduce one state aspect for its current content. 
The human might also be unaware of an exact Sushi recipe. For every recipe an additional state aspect is used to represent possible deviations from the true recipe (e.g.\ different fish type or different rice preparation). 
With 10 locations (with 4 to 37 different possible contents) and 6 recipes we have a large combinatorical size of the state space of $\vert S\vert=27^5 \cdot 21 \cdot 15 \cdot 8^2 \cdot 6* 5^4 \cdot 4^2 \approx 8.4\cdot 10^{16}$. 

\paragraph{Robot task policy}
The robot acts towards the common goal reacting to the human without explicit coordination. It plans the required actions backwards from the current goal. If multiple paths are possible, the robot picks one randomly. So if the human behaves passively, the robot takes the initiative and starts preparation (though at some point, ingredients from human storage might be necessary). 
Otherwise the robot adapts to the human actions. 
The robot's planning module is further used to rate the final states of communication planning. 

\begin{figure}
\centering

\begin{minipage}{.4\linewidth}
\includegraphics[width=\linewidth]{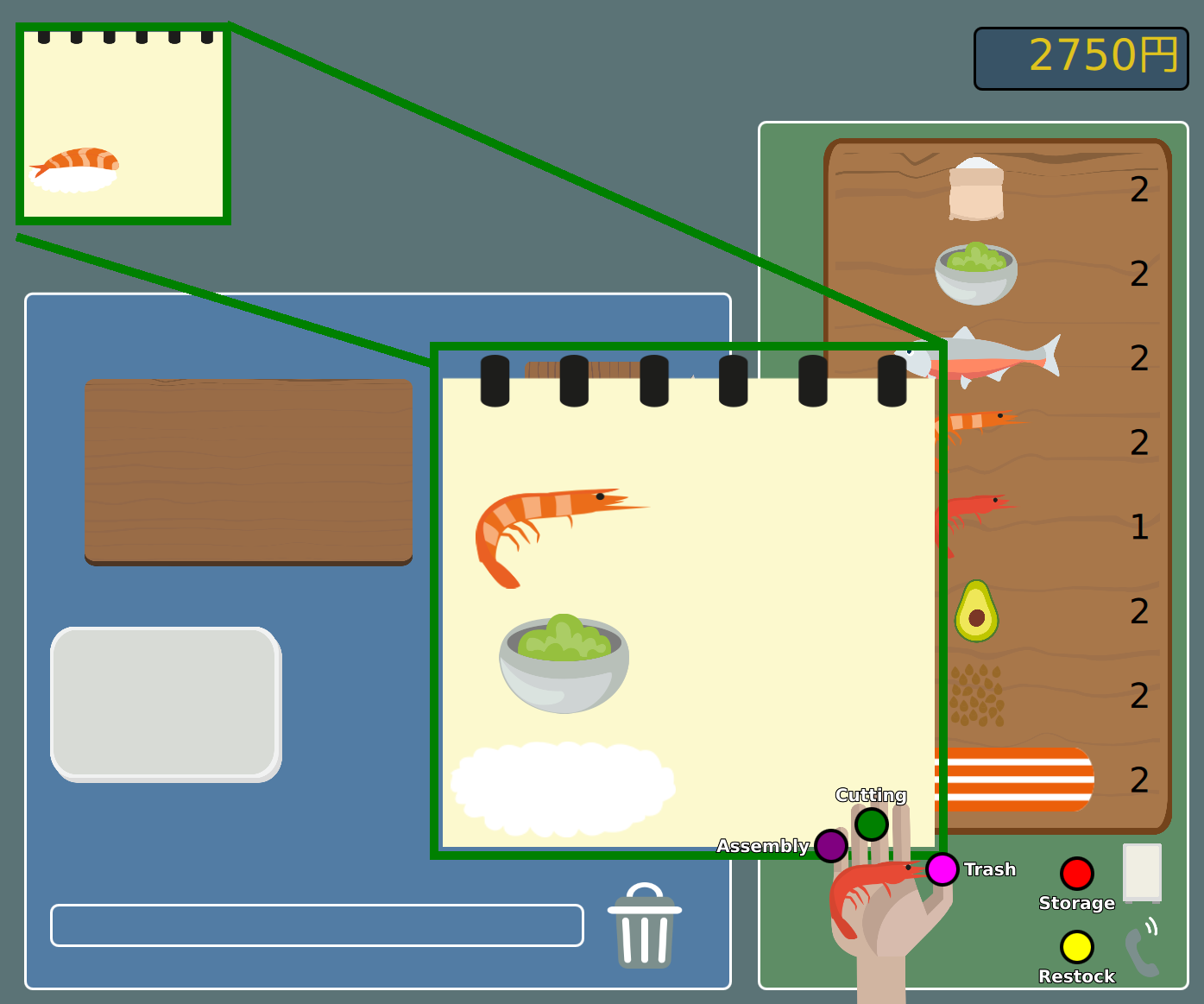}
\end{minipage}
\begin{minipage}{.42\linewidth}
\includegraphics[width=\linewidth]{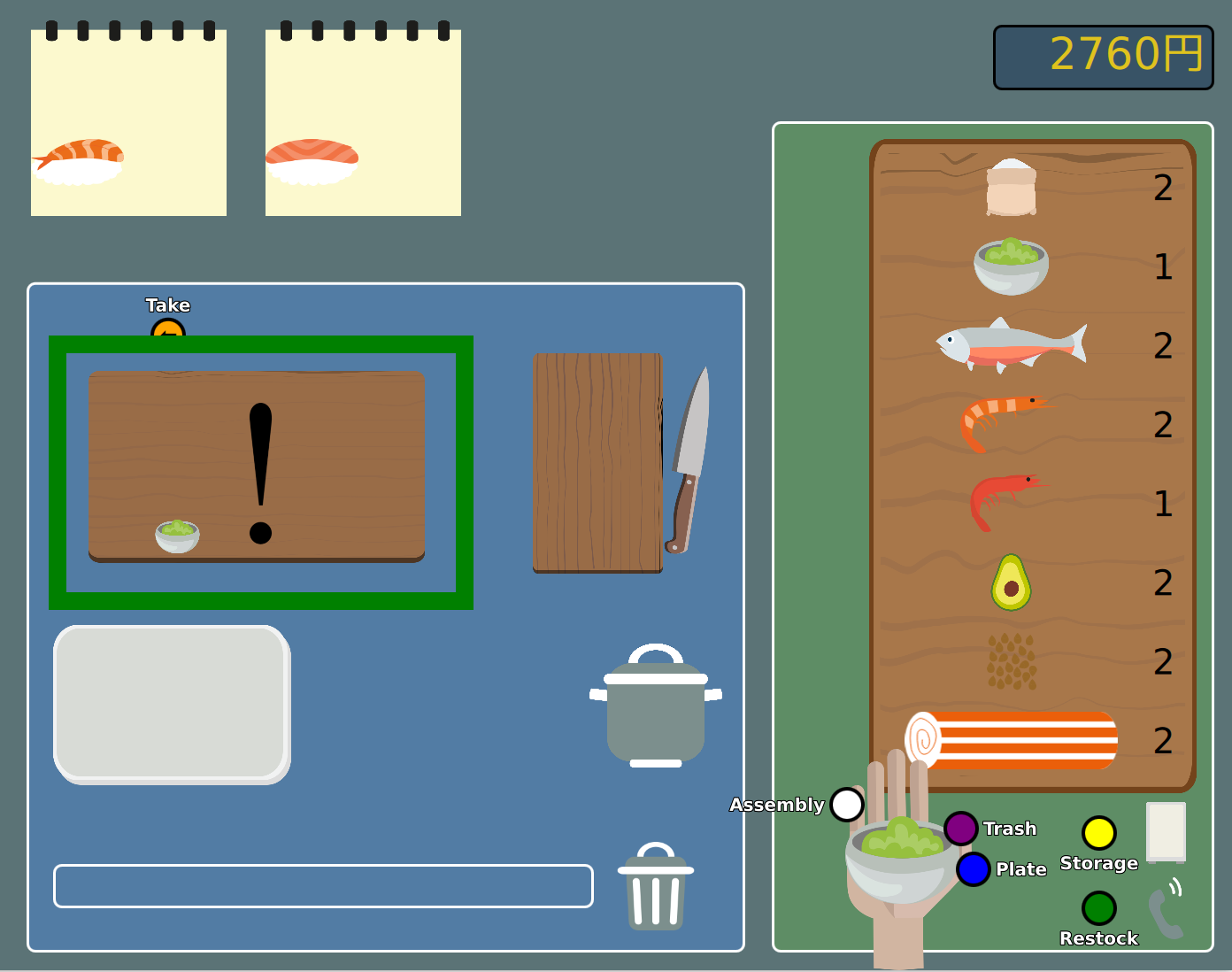}
\end{minipage}

\caption{Communication signals: Show the current recipe (left) or direct attention to a specific location and corresponding content (right).}
\label{signals}
\end{figure}

\paragraph{Design of communication signals}
To illustrate human awareness problems, consider a typical situation. The human has access to two different shellfish types. If the human wrongly believes that the current recipe requires the second instead of the first, this false belief can lead to a longer sequence of bad human actions. The human might take the wrong shellfish, cut it on the cutting board and put it to the assembly board. By showing the recipe with the correct ingredients (see Fig.\ \ref{signals} left), the human can recover from the false belief and related unawareness. 

Besides recipe difficulties, a second source of belief related problems results from coordination with the robot. If the robot started with one order while the human is starting with another, the usage of locations will conflict and need to be resolved. Similar on the action level coordination conflicts can evolve e.g.\ that both agents deliver the same ingredient. 
Highlighting locations of possible coordination conflicts (robot hand or assembly board, see Fig. \ref{signals} right) can give the human the possibility to recognize and solve these before more actions have to be reverted. 

For the effects of these communication signals, we model a certain update of the corresponding human belief aspects (recipe and order aspect for showing recipe, location contents for highlighting).

To respect negative effects of communication as distraction and processing time needs, we count a constant cost corresponding to the cost of 1.5 additional task actions (communication should be triggered if a saving of more than one and a half actions is expected).

\paragraph{Wizard of Oz setting}
Due to this large state space, planning different possible human behaviors (which is used in belief inference and communication planning) gets relatively slow. We could achieve the computation of one time step in about 8s (single core computation @ 3.70GHz) while a typical human action occurred every 2 seconds in the trials. In a prestudy, we tested a limitation of action execution speed to a duration of 8 seconds, however the effects on the task and human performance were significant. The participants had more time to evaluate the current situation leading to fewer and less predictable errors which contradicts evaluation requirements. 

We therefore use a ``Wizard of Oz'' setting, where the available communication actions are triggered by a human expert (first author) in the assisted condition. The robot's task action are selected automatically according to its policy. 

The wizard followed the concept of theory of mind based communication, with the same information base (access to system state and gaze measurements). The wizard was trained in advance to detect and evaluate task situations fast enough.

\subsection{User study}
The target of the user study is to evaluate the benefits of the assistive communication on task performance.

We consider the following hypotheses. A communicative assistance concept based on theory of mind improves joint task performance (H1). The decisions of our communication assistant are similar to the wizard decisions (H2). Compared to alternative communication concepts, our ToM-Com assistant supports a human partner more efficiently, leading to fewer interruptions (H3).  

14 participants, well educated and with mostly technical background, cooperated with the robot on our sushi task. Participants were randomly divided into two groups, one started with an assisted trial, the other started unassisted. The duration per participant was 1 hour and limited the total number of actions performed by the agents. 
The participants had no prior experience with the task and started with instructions and a familiarization phase. 
Afterwards, two trials were recorded, one with and one without assistive communication. After each trial, we collected subjective impressions regarding task understanding, difficulty and assistance via a questionnaire. 
The randomized order of the trials is important, since the increasing experience of the participants has a large effect on their performance. 

\subsection{Results}
We designed the task to evaluate communicative assistance according to our human centric ToM-Com concept. 
The task was difficult for the participants and fulfilled our requirement to generate situations, where a human is not aware of some important aspects. Out of 8515 recorded actions, 587 human errors occurred. We classified 481 of them as caused by different types of belief related awareness problems. The remaining errors were probably caused by color mismatch or erroneous button presses. A false belief of one important aspect can induce a longer sequence of errors, e.g.\ by continuing working with wrong ingredients. We could detect 153 of such error sequences.

\paragraph{Wizard similarity}
Before investigating the performance improvement by communicative support, the similarity between the communication decisions of the wizard and those of our assistance system needs to be evaluated (H2). We play back the recorded data to evaluate communication decisions.

In general, the assistant would communicate more often and more proactively. 
For 74\% of the communication actions by the wizard, a similar decision is selected by the assistant, meaning the robot would communicate in nearby time steps. 
We had a look at situations, where the wizard communicated while the assistant does not. Within these, there are false positives of the wizard and situations, where the wizard reacted late to an awareness problem, while the assistant would have interacted earlier. Still there are cases, where the assistant does not intervene. These occur often during short error sequences where it might have intervened if the sequence would have continued, i.e. evidence accumulation was slower than for the wizard.

\paragraph{Performance improvement by human centric assistance}

We hypothesize that human centric assistance can improve the performance of the participant in the cooperative sushi task (H1). 
As performance measures we consider primarily the number of errors of the human and the length of error sequences. The results are similar for other performance measures such as the time required to complete an order or the number of additional actions.

\begin{figure}
\centering
\begin{minipage}{.46\linewidth}
\includegraphics[width=\linewidth]{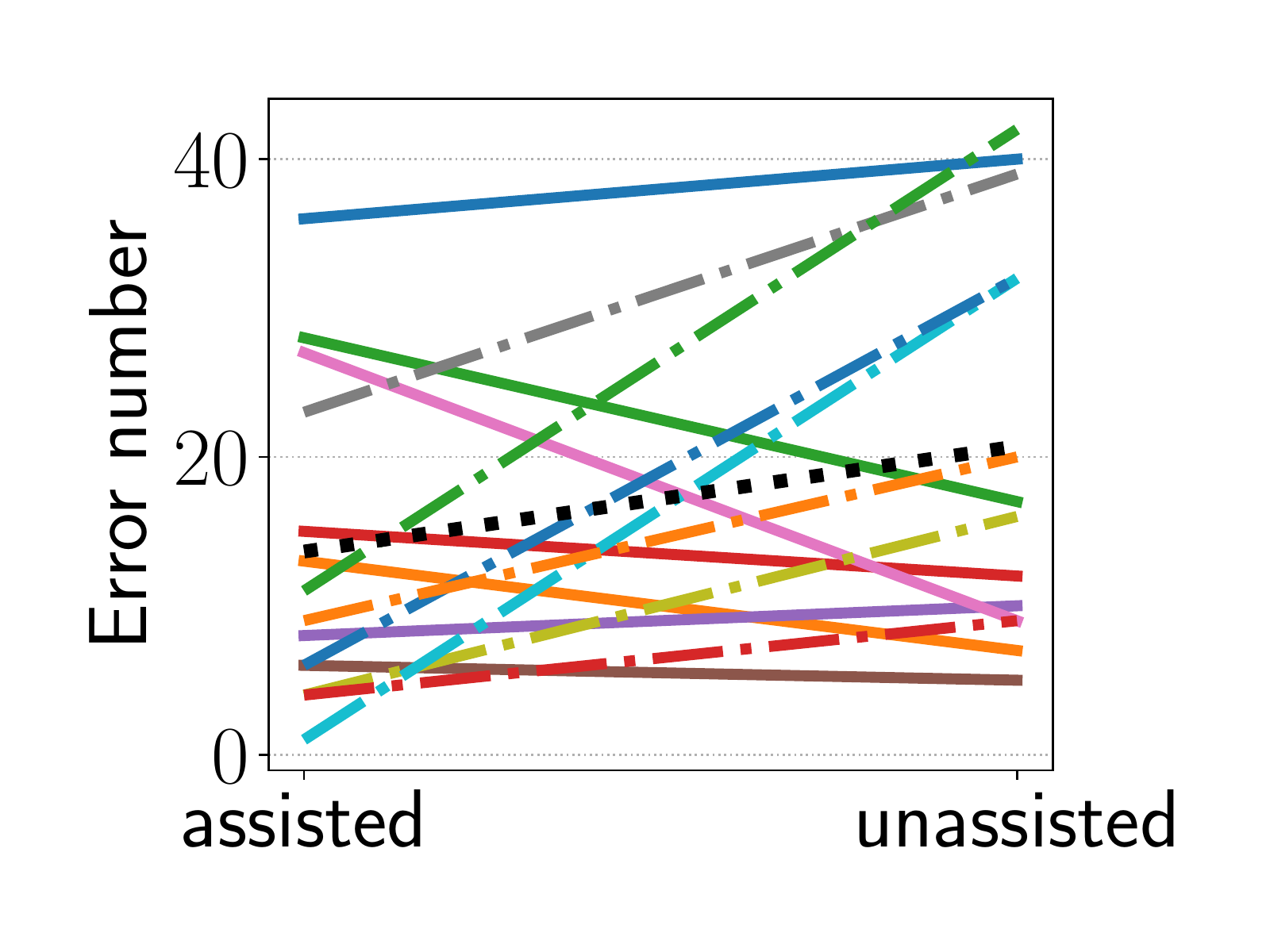}
\end{minipage}%
\begin{minipage}{.54 \linewidth}
\includegraphics[width=\linewidth]{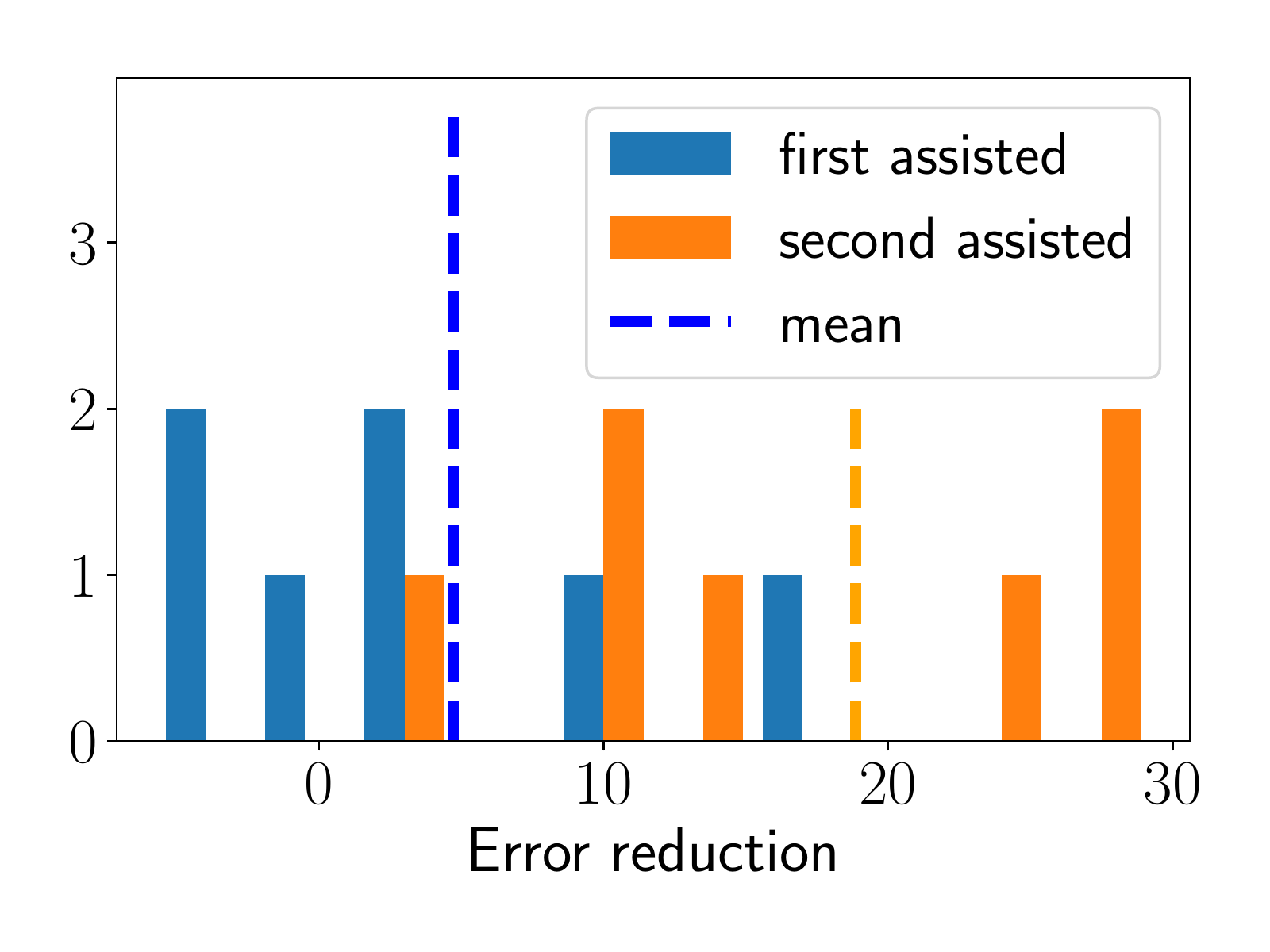}
\end{minipage}

\begin{minipage}{.42\linewidth}
\includegraphics[width=\linewidth]{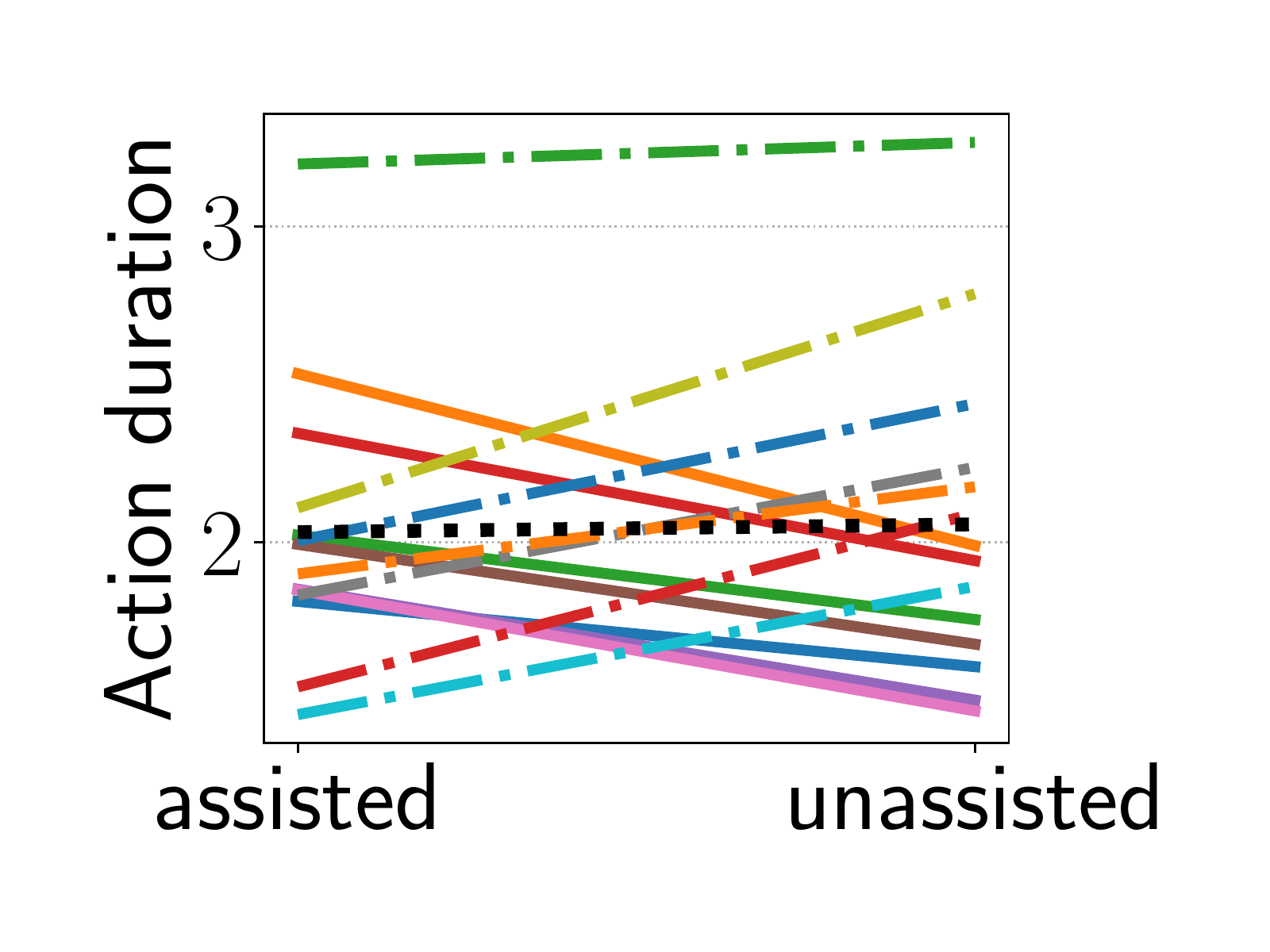}
\end{minipage}%
\begin{minipage}{.48\linewidth}
\includegraphics[width=\linewidth]{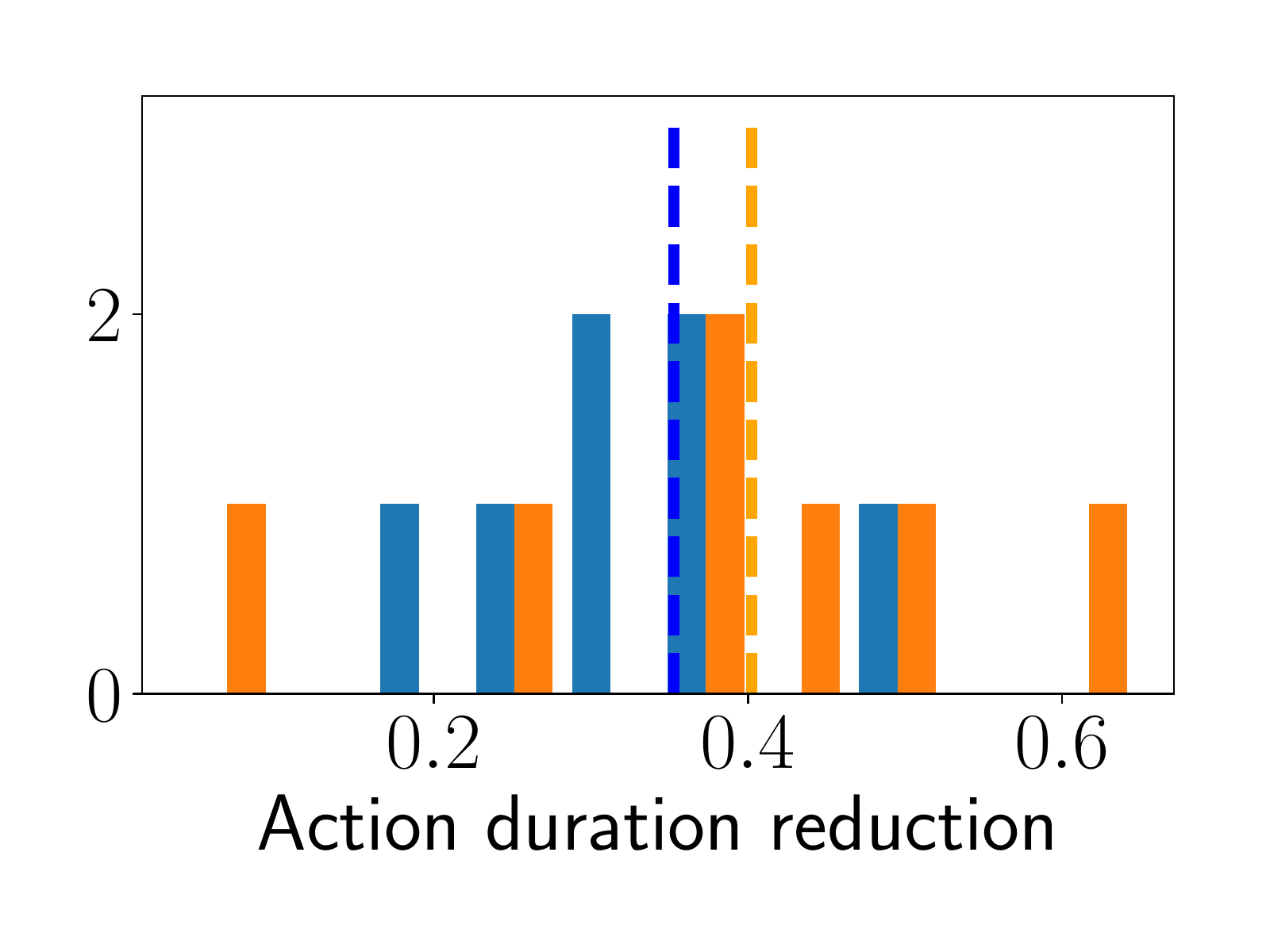}
\end{minipage}
\caption{Error number over assistance condition for each participant, mean in black (top left). To separate assistance from experience effects (action duration at bottom), the top right shows the difference of first vs second trial. The difference in means show the performance improvement due to communicative assistance.}
\label{error_count}
\end{figure}
The number of errors of the participants is shown in Fig. \ref{error_count} top left for the two conditions with and without support. 
Even though we provided time for familiarization with the task, we still found a significant improvement in performance over the sessions. 

To distinguish effects of participants experience and assistance, we assume these to be independent. This is supported by a look at typical action execution duration as a measure for experience, which decreases with experience but seems to be independent of the assistance condition (see Fig. \ref{error_count} bottom). 
To separate the effects, we calculate the performance improvements from the first to the second trial (see Fig. \ref{error_count}, top right). We find a difference between the groups due to the communication condition. The group starting with an assisted followed by an unassisted session shows smaller and even negative changes in performance. 
The difference in means shows the benefits of communication for the joint performance. 
A mixed ANOVA, with assistance condition as within subjects variable and the group as in between variable, shows a significant influence of assistance aspect, $p=0.014$. 
The Pearson correlation factor between assistance condition and error reduction is $r=0.64$ ($p=0.014$).

\begin{figure}
\centering
\includegraphics[width=.9\linewidth]{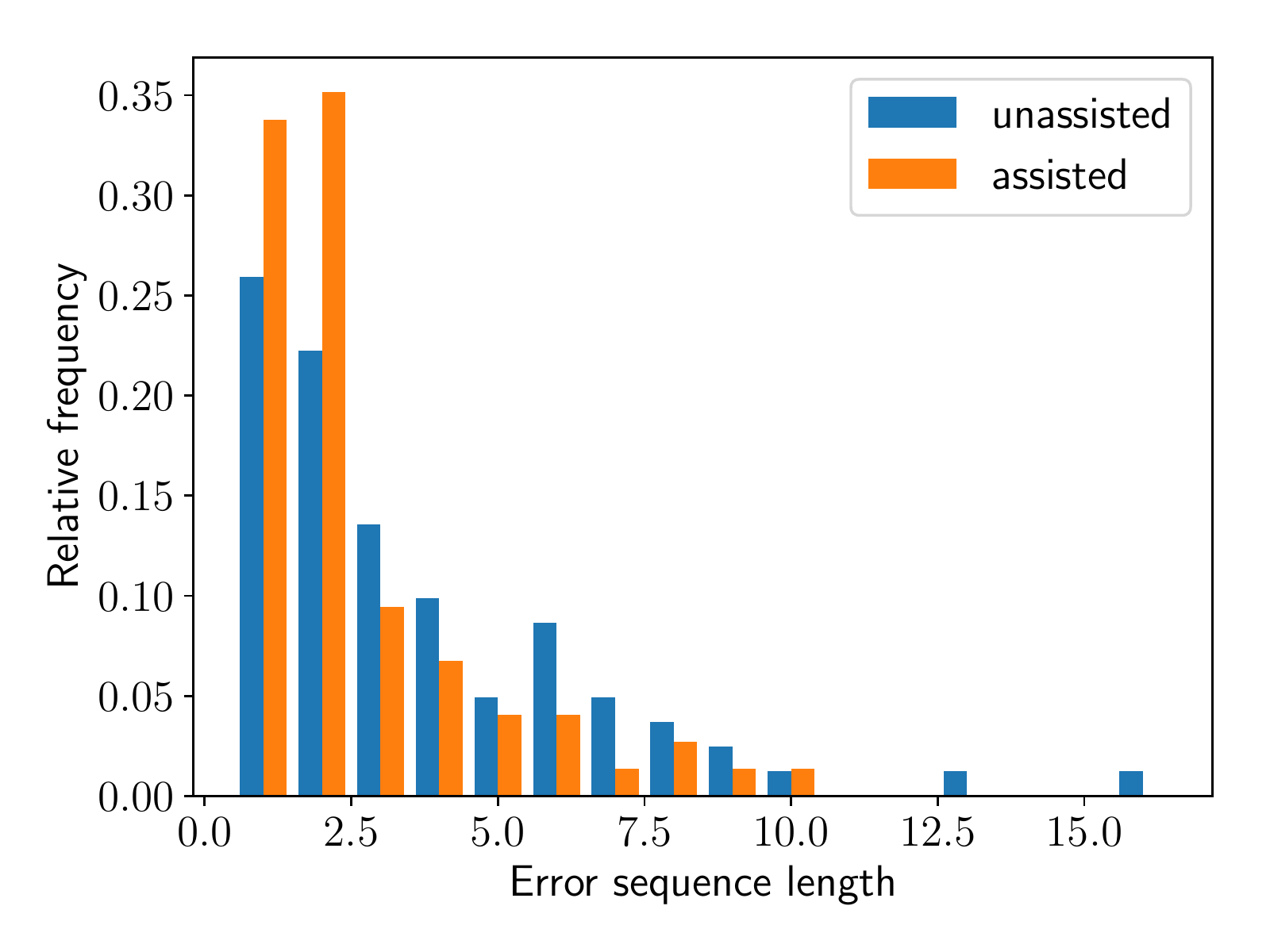}
\caption{Effect of communicative assistance on error sequence lengths (number of related errors). Communication helps the human to recover from false belief situations and leads to shorter error sequences.}
\label{seqlengths}
\end{figure}
The performance improvement due to assistance is further visible when looking at the sequence lengths. In Fig. \ref{seqlengths}, the lengths distribution of error sequences for the assisted and unassisted trials are shown. With communicative support, participants could recover faster from false belief situations, leading to reduced error sequence lengths. 

Here we can also see the importance of understanding the causes of human errors. In the condition without assistance, there are many short error sequences where the human would not profit from disturbing communication.

In some cases, the information actions could seemingly not achieve the intended effects, as there are still longer sequences in the assisted condition. 
We confirmed through the questionnaires, that some participants indeed had difficulties to understand the communication content for some actions.

\paragraph{Acceptance}
We hypothesize that the receiver centric communication approach with sparse interruptions leads to high acceptance by the human participants. 
This is supported by the subjective post trial ratings ("The signals were given too often", "The signals annoyed me") where 7 reported high acceptance, 3 medium acceptance and 3 were undecided (1 did not report acceptance). Further comparison studies may confirm these results and also investigate the relation to understanding and the design of communication. 

\paragraph{Concept comparisons}

Besides the performance improvement related to the condition without communication, we compare the communication decisions to a state of the art assistive communication concept and a Theory of Mind based concept without communication planning. We use the trials from the condition without assistance. 

Typical human centric assistance approaches intervene ``when'' the human deviates from an expected behavior (human error), e.g. by warning or by proposing a good next action. 
This can help the human making good actions at time but might not support the human to understand the situation and the error they made. For the analysis, we assume that participants would follow the action proposal and that this deviation based concept (DEV) would prevent a following human error. 

The ToM based alternative without communication Planning skips the evaluation of communication effects. Instead it decides to communicate when a false or uncertain human belief is estimated, i.e. similar to our previous work \cite{Buehler2018}. It should prevent human errors similar to ToM-Com, but may lead to unnecessary communication, when the false belief aspect is not relevant for current decisions. Here we consider a decision threshold for the expected belief deviation.

In the previous paragraphs, we could show a significant reduction of error sequence length. Therefore we assume, that providing missing information prevents the subsequent errors in the current sequence for both ToM concepts with (ToM-Com) and without (ToM) communication planning.

\begin{figure}
\centering
\includegraphics[width=.85\linewidth]{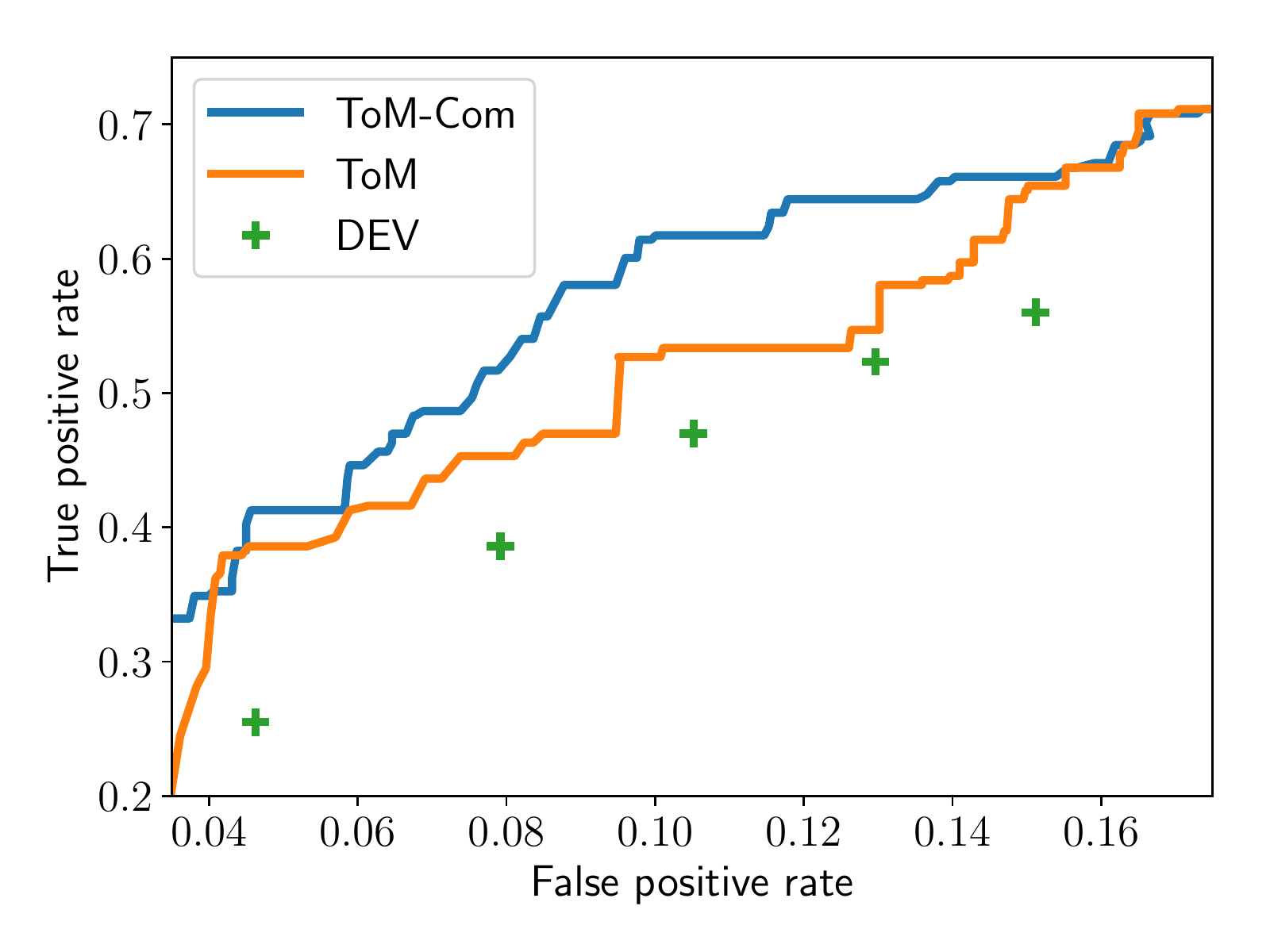}

\caption{ROC curves for communication concepts. Our concept (ToM-Com) with communication costs between 0.5 and 2.4, a theory of mind approach without communication planning (ToM) with erroneous belief thresholds between 0.4 and 0.99, and a simple deviation based approach (DEV) proposing the next 1 to 5 actions after a human error.} 
\label{roc}
\end{figure}
For ToM-Com, the cost of communication is the explicit parameter to handle the false positives against false negatives. The resulting receiver operating characteristics are shown in Fig.\ \ref{roc}. 

For the deviation based communication we vary the number of next actions included in the proposal. 
Our assistant performs clearly better than the compared concepts, providing the chance to prevent more errors while disturbing the human partner less often.

\section{Conclusion}
We evaluated the performance benefits of our human centric communication concept. Based on an understanding of the human behavior through the estimation of the human belief, the relevance of different types of information is evaluated leading to the decision when and what type of information to share to support a human partner. 
At our challenging Sushi task, we could show significant performance improvements compared to the condition without communication in a user study. Further evaluations show benefits compared to a standard approach that intervenes at human errors and proposes good actions as well as to an approach without communication planning. It can prevent more human errors, while avoiding unnecessary disturbances. 
We believe that it is beneficial to enable the human to understand the situation and behave well compared to proposing actions to an unaware human actor. Assistance systems are not perfect, which may lead to confusion if communication is not transparent. We want to treat the causes of human errors and not the symptoms to give the human good decision options. 
We believe that our generic human centric communication concept enables new types of intelligent decision support systems in different application fields. 

\bibliography{ArxivPaper_References}

\end{document}